\pdfoutput=1
\documentclass[11pt]{article}

\usepackage{acl}

\usepackage{times}
\usepackage{latexsym}

\usepackage[T1]{fontenc}
\usepackage[utf8]{inputenc}

\usepackage{microtype}

\usepackage{csquotes}

\usepackage{marvosym}

\usepackage{graphicx}

\usepackage{amssymb}

\usepackage{chngpage}

\usepackage{placeins}

\title{Relevance in Dialogue: Is Less More? An Empirical Comparison of Existing Metrics, and a Novel Simple Metric}

\author{Ian Berlot-Attwell \\
  University of Toronto \\
  Vector Institute \\
  \texttt{ianberlot@cs.toronto.edu} \\\And
  Frank Rudzicz \\
  University of Toronto \\
  Vector Institute \\
  Unity Health Toronto \\
  \texttt{frank@cs.toronto.edu} \\}

\begin{document}
\maketitle
\begin{abstract}
In this work, we evaluate various existing dialogue relevance metrics, find strong dependency on the dataset, often with poor correlation with human scores of relevance, and propose modifications to reduce data requirements and domain sensitivity while improving correlation. Our proposed metric achieves state-of-the-art performance on the HUMOD dataset \cite{humod} while reducing measured sensitivity to dataset by $37$\%-$66\%$. We achieve this without fine-tuning a pretrained language model, and using only $3,750$ unannotated human dialogues and a single negative example. Despite these limitations, we demonstrate competitive performance on four datasets from different domains. Our code, including our metric and experiments, is open sourced\footnote{\url{https://github.com/ikb-a/idk-dialogue-relevance}}.
\end{abstract}

\section{Introduction}

The automatic evaluation of generative dialogue systems remains an important open problem, with potential applications from tourism \cite{tourism} to medicine \cite{covid}. In recent years, there has been increased focus on interpretable approaches \cite{Deriu2021, dstc10} often through combining various sub-metrics, each for a specific aspect of dialogue  \cite{ling_feat, phy-etal-2020-deconstruct, mehri-eskenazi-2020-usr}. One of these key aspects is \enquote{relevance} (sometimes called \enquote{context coherence}), commonly defined as whether \enquote{[r]esponses are on-topic with the immediate dialogue history} \cite{finch-choi-2020-towards}. 

% Xu: Coherence: "In this work, given a dialogue history, we regard as a coherent response an utterance that is thematically correlated and naturally continuing from the previous turns, as well as lexically diverse."

% USR: Maintains Context: Does the response serve as a valid continuation of the preceding conversation?

% Pang: context coherence of a dialogue: the meaningfulness of a response within the context of prior query

% FED: \say{Is the response relevant to the conversation?} and \say{Is the response correct or was there a misunderstanding of the conversation? }

% GRADE: dialogue Coherence, what makes dialogue utterances unified rather than a random group of sentences

% DynaEval: dialogue coherence: considers whether a piece of text is in a consistent and logical manner, as opposed to a random collection of sentences.

%Related to both dialogue quality and relevance is the task of measuring discourse coherence, defined by \citet{zhang-etal-2021-dynaeval} as \enquote{whether a piece of text is in a consistent and logical manner, as opposed to a random 	collection of sentences}. Much like overall dialogue quality, it is clear that, relevance over all turns is necessary, but insufficient, for dialogue coherence. In future, it is possible that coherence metrics such as GRADE \cite{huang-etal-2020-grade} and DynaEval \cite{zhang-etal-2021-dynaeval} may be adapted for solely relevance.

These interpretable approaches have motivated measures of dialogue relevance that are not reliant on expensive human annotations. Such measures have appeared in many recent papers on dialogue evaluation, including USR \cite{mehri-eskenazi-2020-usr}, USL-H \cite{phy-etal-2020-deconstruct}, and others \cite{pang-etal-2020-towards, humod}. Additionally, dialogue relevance has been used directly in training dialogue models \cite{xu-etal-2018-better}.

Despite this work, comparison between these approaches has been limited. Aggravating this problem is that authors often collect human annotations on their own datasets with varying amounts and types of non-human responses. % and, as a result, 
Consequently, %comparing between approaches has been difficult, if not impossible. 
direct comparisons are not possible. It is known that metrics of dialogue \textit{quality} often perform poorly on new test sets of quality ratings \cite{yeh-etal-2021-comprehensive}, but it remains an open question whether poor generalization also plagues the much simpler dialogue relevance task. We address this problem by evaluating and comparing six prior approaches on four publicly available datasets of dialogue annotated with human ratings of relevance. We find poor correlation with human ratings across various methods, with high sensitivity to dataset. 

Based on our observations, we propose a simple metric of logistic regression trained on pretrained BERT NSP features \cite{devlin-etal-2019-bert}, using \enquote{i don't know.} as the only negative example. With this metric, %described below, 
we achieve state-of-the-art correlation on the HUMOD dataset \cite{humod}. %We make our code, data processing, and empirical setup publicly available to encourage more comparable results in future research.
We release our metric and evaluation code to encourage comparable results in future research. 

Our primary contributions are: (i) empirical evidence that current dialogue relevance metrics for English are sensitive to dataset, and often have poor correlation with human ratings, (ii) a simple relevance metric that exhibits good correlation and reduced domain sensitivity, and (iii) the counter-intuitive result that a single negative example can be equally effective as random negative sampling.

\section{Prior metrics}

Prior metrics of relevance in dialogue can generally be divided into more traditional approaches that are token-based, and more current approaches based on large pretrained models. These metrics are given the \textit{context} (i.e., the two-person conversation up to a given point in time), as well as a \textit{response} (i.e., the next speaker's response, also known as the `next turn' in the conversation). From these, they produce a measure of the response's relevance to the context. The ground-truth response (i.e., the `gold response') may or may not be available.

\subsection{$n$-gram approaches}

There have been attempts to use metrics based on $n$-grams from machine-translation and summarization, such as BLEU \cite{papineni-etal-2002-bleu}, ROUGE \cite{lin-2004-rouge}, and METEOR \cite{banerjee-lavie-2005-meteor} in dialogue. However, we discard these approaches due to their limitations: %these approaches are limited: 
they require a ground-truth response, 
and correlate poorly with dialogue %quality \cite{liu-etal-2016-evaluate} and 
relevance \cite{humod}.
%they perform poorly when measuring dialogue quality \cite{liu-etal-2016-evaluate}, and they have poor correlation with human relevance scores \cite{humod}.

\subsection{Average-embedding cosine similarity}

\citet{xu-etal-2018-better} proposed to measure the cosine similarity of a vector representation of the context and the response. Specifically, the context and response are represented via an aggregate (typically an average) of the uncontextualized word embeddings.  This approach can be modified to exploit language models by instead using contextualized word embeddings.

\subsection{Fine-tuned embedding model for Next Utterance Prediction (NUP)}

 This family of approaches combines a word embedding model (typically max- or average-pooled BERT word embeddings) with a simple 1-3 layer MLP, trained for next utterance prediction (typically using negative sampling) \cite{mehri-eskenazi-2020-usr, phy-etal-2020-deconstruct}. The embedding model is then fine-tuned to the domain of interest. In some variants, the model is provided with information in addition to the context and response; e.g., \citet{mehri-eskenazi-2020-usr} %measured relevance on annotated Topical-Chat data \cite{topicalchat} by appending the 
 appended a topic string to the context. This %general architecture and training paradigm have 
 approach has also been directly used as a metric of overall dialogue quality \cite{ghazarian-etal-2019-better}. In this paper, we focus on the specific implementation by \citet{phy-etal-2020-deconstruct}%. They use max-pooled BERT embeddings that are passed into a single-layer MLP followed by softmax with two classes. Binary cross-entropy loss and random sampling of negative examples is used at train time. 
 : max-pooled BERT embeddings passed into a single-layer MLP followed by two-class softmax, trained with binary cross-entropy (BCE) loss and random sampling of negative samples.
 
 %Note that, for methods that are fine-tuned or otherwise require training, it will often be the case that annotated relevance data is not available on the domain of interest. As a result, the model performance (i.e., correlation with human annotations) cannot be measured on a validation set, and some other means must be used to determine when training must stop (e.g., loss on the surrogate task, or halting after a certain number of epochs). It is therefore important that either the surrogate loss correlates well with the model performance, or the true validation curves of these methods be relatively smooth and monotone so as to reduce the risk of halting training on a model with poor performance. 
 
 Note that, for methods that are fine-tuned or otherwise require training, it will often be the case that annotated relevance data is not available on the domain of interest. As a result, model performance cannot be measured on a validation set during training. Therefore, either the method must be trained to convergence on the training set, or a different method other than validation set performance must be employed to reduce the risk of halting training on a model with poor performance. 
 
 Another concern with using trained metrics to evaluate trained dialogue systems is that they may both learn the same patterns in the training data. An extreme example would be a dialogue model that learns only to reproduce responses from the training data verbatim, and a relevance metric that learns to only accept verbatim responses from the training data. We believe that this risk can be reduced by training the metric on separate data from the model. However, this approach is only practical if the metric can be trained with a relatively small amount of data and therefore does not compete with the dialogue model for training examples. Alternatively, a sufficiently generalizable metric may be trained on data from a different domain.

\subsection{Normalized conditional probability}

\citet{pang-etal-2020-towards} also exploited pretrained models, however they instead relied on a generative language model (specifically GPT-2). Their proposed metric is the conditional log-probability of the response given the context, normalized to the range $[0,1]$ (see Appendix \ref{app:pang} for details). %Specifically, for a context $q$ with candidate response $r$, their proposed relevance score is defined as: $c(q\,|\,r) = - \frac{\max(c_{5th}, \frac{1}{|r|} \log P(r\,|\,q)) - c_{5th}}{c_{5th}}$, where $|r|$ is the number of tokens in the response, $P(r\,|\,q)$ is the conditional probability of the response given the context under the language model, and $c_{5th}$ is the $5^{th}$ percentile of the distribution of $\frac{1}{|r|} \log P(r\,|\,q)$ over the examples being evaluated.

\citet{mehri-eskenazi-2020-unsupervised} also relied on a generative language model (specifically,  DialoGPT \cite{zhang-etal-2020-dialogpt}), however their approach measured the probability of followup-utterances, e.g., \enquote{Why are you changing the topic?} to indicate irrelevance. Their relevance and correctness scores are defined as
$c(q\,|\,r) = -\sum_{i=1}^{|n|} \log P(n_i \,|\, r, q)$,
where $n_i \in n$ is a negative response suggesting irrelevance or incorrectness. Note that positive utterances can be used, however the author's measures of correctness and relevance only used negative utterances.

\section{Datasets used for analysis}

\begin{table*}[ht]
\scriptsize
\begin{adjustwidth}{-2cm}{-2cm}
\centering
	\begin{tabular}{|l|p{3cm}|r|p{1cm}|p{1cm}|p{3cm}|p{2cm}|}
	\hline
	Dataset &  Superset & Contexts &  Turns per Context &  Responses per \newline Context &   Response types & {Relevance\newline Annotation} \\
	\hline 
	\hline 
	HUMOD \cite{humod} & Cornell movie dialogue \cite{danescu-niculescu-mizil-lee-2011-chameleons} & $4,750$ & $2$-$7$ & $2$ & Human, Random Human & Likert $1$-$5$ \\
	\hline 
	USR-TC \cite{mehri-eskenazi-2020-usr} & Topical Chat \cite{topicalchat} & $60$ & $1$-$19$ & $6$ & Human (x2), Transformer (x4) & Likert $1$-$3$ \\
	\hline 
	P-DD \cite{pang-etal-2020-towards} & DailyDialogue \cite{li-etal-2017-dailydialog} & $200$ & $1$ & $1$ & LSTM & Likert $1$-$5$ \\
	\hline 
	FED \cite{mehri-eskenazi-2020-unsupervised} & N/A & $375$ & $3$-$33$ & $1$ & Human, Meena \cite{meena}, or Mitsuku & Likert $1$-$3$\newline (relevance and \newline correctness) \\
	\hline 
	\end{tabular}
\end{adjustwidth}
\caption{Summary of datasets used.}
\label{tab:datasets}
\end{table*}

A literature review reveals that many of these methods have never been evaluated on the same datasets. As such, it is unclear both how these approaches compare, and how well they generalize to new data. For this reason, we consider four publicly available English datasets of both human and synthetic dialogue with human relevance annotations. All datasets are annotated with Likert ratings of relevance from multiple reviewers; following \citet{humod}, we average these ratings over all reviewers. Due to variations in data collection procedures, as well as anchoring effects \cite{acute}, Likert ratings from different datasets may not be directly comparable. %For this reason, 
Consequently, we keep the datasets separate. This %has the additional benefit of allowing 
also allows us to observe generalization across datasets. 

Altogether, our selected datasets cover a wide variety of responses, including human, LSTM, Transformer, Meena \cite{meena}, and Mitsuku\footnote{2019 Loebner prize winning system} generated responses, and random distractors. See Table \ref{tab:datasets} for an overview.

\subsection{HUMOD Dataset}

The HUMOD dataset \cite{humod} is an annotated subset of the Cornell movie dialogue dataset  \cite{danescu-niculescu-mizil-lee-2011-chameleons}. The Cornell dataset consists of $220,579$ conversations from $617$ films. The HUMOD dataset is a subset of $4,750$ contexts, each consisting of between two and seven turns. Every context is paired with both the original human response, and a randomly sampled human response. Each response is annotated with crowd-sourced ratings of relevance from 1-5. The authors measured inter-annotator agreement via Cohen's kappa score \cite{cohen1968weighted}, and it was found to be 0.86 between the closest ratings, and 0.42 between randomly selected ratings. Following the authors, we split the dataset into a training set consisting of the first $3,750$ contexts, a validation set of the next $500$ contexts, and a test-set of the remaining $500$ contexts. As it is unclear how HUMOD was subsampled from the Cornell movie dialogue dataset, we do not use the Cornell movie dialogue dataset as training data. %for any of our methods.

\subsection{USR Topical-Chat Dataset (USR-TC)}

The USR-TC dataset is a subset of the Topical-Chat (TC) dialogue dataset \cite{topicalchat} created by \citet{mehri-eskenazi-2020-usr}. The Topical-Chat dataset consists of approximately $11,000$ conversations between Amazon Mechanical Turk workers, each grounding their conversation in a provided reading set. The USR-TC dataset consists of 60 contexts taken from the TC frequent test set, each consisting of 1-19 turns. Every context is paired with six responses: the original human response, a newly created human response, and four samples taken from a Transformer dialog model \cite{attentionisallyouneed}. Each sample follows a different decoding strategy, namely: argmax sampling, and nucleus sampling \cite{nucleussample} at the rates $p=0.3,0.5,0.7$, respectively. Each response is annotated with a human 1-3 score of relevance, produced by one of six dialogue researchers. The authors reported an inter-annotator agreement of 0.56 (Spearman's correlation). We divide the dataset evenly into a validation and test set, each containing 30 contexts. We use the TC train set as the training set.

\subsection{\citet{pang-etal-2020-towards} Annotated DailyDialogue Dataset (P-DD)}

The P-DD dataset \cite{pang-etal-2020-towards} is a subset of the DailyDialogue (DD) dataset \cite{li-etal-2017-dailydialog}. The DailyDialogue dataset consists of $13,118$ conversations scraped from %various websites, specifically digital spaces 
websites where English language learners could practice English conversation.  The P-DD dataset contains 200 contexts, each %consisting 
of a single turn %. Each context is 
and paired with a single synthetic response, generated by a 2-layer LSTM \cite{lstm}. Responses are sampled using top-K sampling for $k\in \{1, 10, 100 \}$; note that $k$ varies by context. Each response is annotated with ten crowdsourced 1-5 ratings of relevance %. The authors reported that 
with a reported inter-annotator Spearman's correlation % varied 
between 0.57 and 0.87. Due to the very small size of the dataset (only 200 dialogues in total), and the lack of information on how the contexts were sampled, we %choose to 
use this dataset exclusively for testing.

\subsection{FED Dataset}

The FED dataset \cite{mehri-eskenazi-2020-unsupervised}, consists of 375 annotated dialogue turns taken from 40 human-human, 40 human-Meena \cite{meena}, and 40 human-Mitsuku conversations. We use a subset of the annotations, specifically turnwise relevance, and turnwise correctness (the latter defined by the authors as whether there was a \enquote{a misunderstanding of the conversation}). As the authors note, their definition of correctness is often encapsulated within relevance; we thus evaluate on both annotations. Due to the small size, we used this dataset only for testing.

\section{Evaluating Prior Metrics}

\begin{table*}[ht]
\scriptsize
\begin{adjustwidth}{-2cm}{-2cm}
\centering
	\begin{tabular}{|l|rr|rr|rr|rr|rr|}
	\hline
	 & \multicolumn{2}{|c|}{HUMOD} & \multicolumn{2}{|c|}{USR-TC} & \multicolumn{2}{|c|}{P-DD} & \multicolumn{2}{|c|}{FED-Correctness} & \multicolumn{2}{|c|}{FED-Relevance}\\
	\hline
	\textbf{Prior Metric} & S  & P  & S & P & S & P & S & P & S & P\\
	\hline 
	% Note: Originally ran without NLTK tokenizer for P-DD
	%COS-FT          & $ 0.09$ & $ 0.10$  & *$ 0.26$ & *$0.24$ & $ -0.10$ & $ -0.11$ \\
	COS-FT          & $ 0.09$ & $ 0.10$  & *$ 0.26$ & *$0.24$ & $ -0.02$ & $ -0.04$ & $0.08$ & $0.04$  & $0.11$ & $0.07$ \\
	COS-MAX-BERT    & *$ 0.13$ & *$ 0.10$  & *$ 0.20$ & $ 0.14$ & $ 0.03$ & $ 0.02$ & $0.03$ & $0.01 $  & $0.06$ & $0.04$ \\
	COS-NSP-BERT    & $ 0.08$ & $ 0.06$   &  $ 0.08$ & $ 0.09$ & *$ 0.30$ & *$ 0.23$ & $-0.03$ & $-0.01$  & $-0.04$ & $-0.02$ \\
	NORM-PROB       & *$ 0.19$ & *$ 0.16$  & *$ -0.24$ & *$ -0.26$ & *$ 0.65$ & *$ 0.59$ & $0.05$ & $0.06$  & $0.07$ & $0.07$ \\ 
	FED-CORRECT          & $-0.06$ & $-0.04$  & $-0.08$ & $-0.12$ & *$-0.25$ & *$-0.26$ & *$0.17$ & *$0.17$  & *$0.15$ & *$0.15$ \\
	FED-RELEVANT         & $-0.06$ & $-0.05$  & $-0.08$ & $-0.12$ & *$-0.26$ & *$-0.27$ & *$0.17$ & *$0.17$  & *$0.15$ & *$0.15$ \\ 
	\hline 
	%\vspace{1mm}
	GRADE              & *0.61       & *0.61       & 0.00         & 0.03         & *0.70         & *0.68         & 0.12    & 0.12    & *0.15    & *0.15    \\
	DYNA-EVAL & *0.09       & *0.10       & 0.10         & 0.10         & 0.00         & -0.02         & *0.26    & *0.27    & *0.32    & *0.31    \\
	\hline
	% Corrected: originally ran without NLTK tokenizer for P-DD (although, shouldn't have much any impact)
	%NUP-BERT (H)    & *0.33 (0.02) & *0.37 (0.02)  & 0.10 (0.02)  & *0.22 (0.01)  & *0.62 (0.03) & *0.54 (0.02) \\
	%NUP-BERT (TC-S) & *0.29 (0.02) & *0.35 (0.03) & \Ankh 0.17 (0.03) & \Ankh 0.20 (0.04)  & *0.58 (0.05) & *0.56 (0.04) \\
	%NUP-BERT (TC)   & *0.30 (0.01) & *0.38 (0.00)   & 0.16 (0.02)  & *0.21 (0.02)& *0.62 (0.04) & *0.58 (0.03) \\
	NUP-BERT (H)    & *0.33 (0.02) & *0.37 (0.02)  & 0.10 (0.02)  & *0.22 (0.01)  & *0.62 (0.04) & *0.54 (0.02) & \Ankh 0.14 (0.04) & *0.21 (0.03) & *0.22 (0.01) & *0.30 (0.01) \\
	NUP-BERT (TC-S) & *0.29 (0.02) & *0.35 (0.03) & \Ankh 0.17 (0.03) & \Ankh 0.20 (0.04)  & *0.58 (0.05) & *0.56 (0.04) & 0.05 (0.04)  & 0.12 (0.01)  & \Ankh 0.16 (0.04) & *0.21 (0.01) \\
	NUP-BERT (TC)   & *0.30 (0.01) & *0.38 (0.00)   & 0.16 (0.02)  & *0.21 (0.02)& *0.62 (0.05) & *0.58 (0.04) & 0.06 (0.01)  & \Ankh 0.12 (0.02) & *0.18 (0.02) & *0.23 (0.01) \\
	\hline 
	\end{tabular}
\end{adjustwidth}
\caption{Spearman (S) and Pearson (P) correlations of baseline models with average human ratings on the test sets. BERT-NUP is averaged over three runs, with the standard deviation reported in brackets. Training data is specified in brackets: (H) signifies HUMOD, (TC) signifies the Topical Chat training set, and (TC-S) signifies a subset of TC containing $3,750$ dialogues (same size as the HUMOD train set). `*' indicates all trials were significant at the $p<0.01$ level. `\Ankh' indicates at least one trial was significant. Note that most cosine and language-model based metrics attain negative correlation with human scores.}
\label{tab:test_prior}
\end{table*}

\begin{table*}[ht]
%\small
\scriptsize
\begin{adjustwidth}{-2cm}{-2cm}
    \centering
    \begin{tabular}{|l|rr|rr|rr|rr|rr|}
\hline
 & \multicolumn{2}{|c|}{HUMOD} & \multicolumn{2}{|c|}{USR-TC} & \multicolumn{2}{|c|}{P-DD} & \multicolumn{2}{|c|}{FED-Correctness} & \multicolumn{2}{|c|}{FED-Relevance}\\
\hline
\textbf{Prior Metric} & S  & P  & S & P & S & P & S & P & S & P\\
\hline 
NUP-BERT (H)    & *0.33 (0.02) & *0.37 (0.02)  & 0.10 (0.02)  & *0.22 (0.01)  & *\textbf{0.62} (0.04) & *0.54 (0.02) & \Ankh 0.14 (0.04) & *0.21 (0.03) & *0.22 (0.01) & *\textbf{0.30} (0.01) \\
NUP-BERT (TC-S) & *0.29 (0.02) & *0.35 (0.03) & \Ankh 0.17 (0.03) & \Ankh 0.20 (0.04)  & *0.58 (0.05) & *0.56 (0.04)  & 0.05 (0.04)  & 0.12 (0.01)  & \Ankh 0.16 (0.04) & *0.21 (0.01) \\
NUP-BERT (TC)   & *0.30 (0.01) & *0.38 (0.00)   & 0.16 (0.02)  & *0.21 (0.02)& *\textbf{0.62} (0.05) & *\textbf{0.58} (0.04) & 0.06 (0.01)  & \Ankh 0.12 (0.02) & *0.18 (0.02) & *0.23 (0.01) \\
\hline
%BERT NSP & *0.59 & *0.40 & 0.17 & *0.25 & *0.53 & *0.31 & 0.12 & 0.10 & *0.21 & *0.18 \\
 IDK (H)          & *\textbf{0.58} (0.00)  & *\textbf{0.58} (0.00)     & \textbf{0.18} (0.00)  & *\textbf{0.24} (0.00)  & *0.53 (0.00)  & *0.48 (0.01)  & *\textbf{0.15} (0.00) & *\textbf{0.23} (0.00)  & *\textbf{0.24} (0.00) & *0.29 (0.00) \\
 IDK (TC-S)      & *\textbf{0.58} (0.00)  & *\textbf{0.58} (0.00)     & \textbf{0.18} (0.00)  & *0.22 (0.00)  & *0.54 (0.01)  & *0.49 (0.01)  & *\textbf{0.15} (0.00) & *\textbf{0.23} (0.00)  & *\textbf{0.24} (0.00) & *0.29 (0.00) \\
\hline
\end{tabular}
\end{adjustwidth}
    \caption{Comparison of our proposed metric (IDK) against the NUP-BERT baseline on the test set. Note the strong improvement on HUMOD and equivalent, or slightly improved performance on USR-TC, at the cost of performance loss on P-DD. Note IDK (H) and IDK (TC-S) performance is almost identical, suggesting that IDK performance is largely independent of training data. %As P-DD consists only of synthetic data generated from an LSTM, we consider this tradeoff to be acceptable.
    }
    \label{tab:test_idk}
\end{table*}

%\end{landscape}

For each of the aforementioned datasets, we evaluate the following relevance metrics:

\begin{itemize}
\setlength{\itemsep}{-1pt}
    \item COS-FT: average fastText \footnote{\url{https://fasttext.cc/}} embedding cosine similarity. Code by \citet{csaky-etal-2019-improving} %Implementation %\footnote{\url{https://github.com/ricsinaruto/dialog-eval}} provided 
    %by \citet{csaky-etal-2019-improving}.
    \item COS-MAX-BERT: Cosine similarity with max-pooled BERT contextualized word embeddings, inspired by BERT-RUBER \cite{ghazarian-etal-2019-better}
    \item COS-NSP-BERT: Cosine similarity using the pretrained features extracted from the [CLS] token used by next-sentence-prediction head.
    \item NUP-BERT: Fine-tuned BERT next-utterance prediction approach. Implementation %\footnote{\url{https://github.com/vitouphy/usl_dialogue_metric}} provided 
    by \citet{phy-etal-2020-deconstruct}. We experiment with fine-tuning BERT to the HUMOD train set (3750 dialogues), the full TC train set, and TC-S (a subset of the TC training set containing $3,750$ dialogues).
    \item NORM-PROB: GPT-2 based normalized conditional-probability; approach and implementation by \citet{pang-etal-2020-towards}; note that the P-DD dataset was released in the same paper.
    \item FED-RELEVANT \& FED-CORRECT: DialoGPT based normalized conditional-probability; approach and implementation %\footnote{\url{https://github.com/shikib/fed}} 
    by \citet{mehri-eskenazi-2020-unsupervised}%; relevance and correctness sub-metrics respectively.
\end{itemize}

In all cases, we use hugging-face \texttt{bert-base-uncased} as  the pretrained BERT model. Only NUP-BERT was fine-tuned. To prevent an unfair fitting to any specific dialogue model, and to better reflect the evaluation of a new dialogue model, only human responses were used at train time. All hyperparameters were left at their recommended values. NUP-BERT performance is averaged over 3 runs.

%Also note that $n$-gram approaches were not evaluated as there is previous evidence suggesting no correlation \cite{humod}, and these methods require a gold-truth reference which is unavailable on the P-DD and FED datasets.

%We do not evaluate $n$-gram approaches due to prior work suggesting no correlation \cite{humod}, and a lack of gold-truth references.

Note that we also evaluate GRADE \cite{huang-etal-2020-grade} and DYNA-EVAL \cite{zhang-etal-2021-dynaeval}; however these do not measure relevance, but rather \textit{dialogue coherence}: \enquote{whether a piece of text is in a consistent and logical manner, as opposed to a random 	collection of sentences}  \cite{zhang-etal-2021-dynaeval}. As relevance is a major aspect of dialogue coherence, we include these baselines for completeness. As both metrics are graph neural networks intended for larger train sets, we use checkpoints provided by the authors. GRADE is trained on DailyDialogue \cite{li-etal-2017-dailydialog}, and DynaEval on Empathetic Dialogue \cite{rashkin-etal-2019-towards}. Both are trained with negative sampling, with GRADE constructing more challenging negative samples.

A summary of the authors' stated purpose for each metric can be found in the Appendix \ref{app:purpose}.

\subsection{Analysis}

Table \ref{tab:test_prior} makes it %immediately 
clear that the normalized probability 
%(NORM-PROB \& FED) 
and cosine similarity 
%(COS-FT, COS-MAX-BERT, COS-NSP-BERT) 
approaches do not generalize well across datasets. Although NORM-PROB %works very well 
excels on the P-DD dataset, it has weak performance on HUMOD and %has, in fact, 
a significant \textit{negative} correlation on USR-TC. Likewise the FED metrics perform %relatively 
well on the FED data, but are negatively correlated on all other datasets. Consequently, we believe that the NORM-PROB and FED metrics are overfitted to their corresponding datasets.  Similarly, although %the cosine-similarity approach using FastText word embeddings 
COS-FT has the best performance on the USR-TC dataset, it performs poorly on HUMOD, and has negative correlation on P-DD. As such, it is clear that, while both cosine-similarity and normalized probability approaches can perform well, they have serious limitations. They are very sensitive to the domain and models under evaluation, and are capable of becoming negatively correlated with human ratings under suboptimal conditions. 

Looking at the \textit{dialogue coherence} metrics, %we find that 
DYNA-EVAL performs strongly on FED, and weakly on all other datasets. GRADE performs very strongly on HUMOD and P-DD (the latter, likely in part as it was trained on DailyDialogue), %however GRADE 
but is uncorrelated on USR-TC. Given that these metrics were not intended to measure relevance, uneven performance is to be expected as relevance and \textit{dialogue coherence} will not always align.

The final baseline, NUP-BERT, is quite competitive, outperforming each of the other baselines on at least 2 of the datasets. Despite this, we can see that performance on HUMOD, USR-TC, and FED is still fairly weak. We can also observe that NUP-BERT has some sensitivity to the domain of the training data; fine-tuning on HUMOD data results in lower Spearman's correlation on USR-TC, and fine-tuning on USR-TC performs worse on the FED datasets. However, the amount of training data (TC vs TC-S) has little impact. 

Overall, the results of Table \ref{tab:test_prior} are concerning as they suggest that at least five current approaches generalize poorly across either dialogue models or domains. %However, this reinforces our previous conclusion that these approaches are highly sensitive to the dataset. The absolute performance of all metrics studied, including our own, vary considerable by dataset. Furthermore, even the relative performance of closely related metrics such as IDK and NUP-BERT, or COS-FT and COS-NSP-BERT, varies considerably between datasets. 
The absolute performance of all metrics studied vary considerably by dataset, and the relative performance of closely related metrics such as COS-FT and COS-NSP-BERT, or NUP-BERT with different training data, varies considerably between datasets. 
As a result, research into new dialogue relevance metrics is required. Furthermore, it is clear that %the methodology for the evaluation of dialogue relevance metrics 
the area's evaluation methodology  must be updated to use various dialogue models in various different domains. 

\section{IDK: A metric for dialogue relevance}

Based on these results, we propose a number of modifications to the NUP-BERT metric to produce a novel metric that we call IDK (\enquote{I Don't Know}). The %overall 
architecture is mostly unchanged, however the training procedure and the %exact 
features used are altered.

First, based on the observation that the amount of training data has little impact, we %decide to 
freeze BERT features %entirely 
and do not fine-tune to the domain. Additionally, whereas the NUP-BERT baseline uses max-pooled BERT word embeddings, we %instead 
use the pre-trained next sentence prediction (NSP) features: \enquote{(classification token) further processed by a Linear layer and a Tanh activation function [...] trained from the next sentence prediction (classification) objective during pre-training}\footnote{\url{https://huggingface.co/transformers/v2.11.0/model_doc/bert.html}}.

Second, to improve generalization and reduce variation in training (particularly important as the practitioner typically has no annotated relevance data), and operating on the assumption that relevance is captured by a few key dimensions of the NUP features, we add L1 regularization to our regression weights ($\lambda = 1$). Note that %we also experimented with L2 regularization and found similar performance on the validation sets (see Appendix, Table \ref{tab:all_valid}).
experiments with L2 regularization yielded similar validation set performance (see Appendix, Table \ref{tab:all_valid}).

Third, in place of random sampling we use a fixed negative sample, ``i don't know". %, at all time steps. 
This allows us to train the model on less data. 

Additionally, we %perform a minor simplification of 
simplify the model, %to reduce the number of weights, 
using logistic regression in place of 2-class softmax. We train for 2 epochs using BCE loss -- the same as the NUP-BERT baseline. We use the Adam optimizer \cite{adam} with an initial learning rate of $0.001$, and batch size 6.

 Table \ref{tab:test_idk}  reports the correlation between the metric's responses and the average human rating. We achieve a Pearson's correlation on HUMOD of 0.58, surpassing HUMOD baselines \cite{humod}, and achieving parity with GRADE ($0.61$). %This performance is also  close to the supervised baseline of 0.602 using a supervised fine-tuned BERT model \cite{humod}. 
%We have included examples of the our metric's output on the HUMOD dataset as well as scatter plot of IDK vs human scores in Appendices \ref{app:example_eval} and \ref{app_scatter}, respectively.
Examples of the our metric's output on the HUMOD dataset, and a scatter plot of IDK vs human scores are in Appendices \ref{app:example_eval} and \ref{app_scatter}, respectively.

Compared to NUP-BERT, %we see that 
our proposed metric provides strong improvement on the HUMOD dataset and equivalent or stronger performance on USR-TC and FED, at a cost of %reduced 
performance on P-DD. In particular, IDK (TC-S) performance on the FED datasets is considerably stronger than NUP-BERT (TC-S). As the performance drop on P-DD is less than the performance gain on HUMOD, and as HUMOD is human data rather than LSTM data, we consider this tradeoff to be a net benefit. %However, this reinforces our previous conclusion that these approaches are highly sensitive to the dataset. The absolute performance of all metrics studied, including our own, vary considerable by dataset. Furthermore, even the relative performance of closely related metrics such as IDK and NUP-BERT, or COS-FT and COS-NSP-BERT, varies considerably between datasets. 

Compared to GRADE in particular, we have reduced performance on P-DD, equivalent performance on HUMOD, and stronger performance on USR-TC and FED (in particular, correlation on the USR-TC dataset is non-zero). It is worth noting that, in general, our approach does not out-perform the baselines in {\em all} cases -- only the {\em majority} of cases. As such, when annotated human data is not available for testing, it would appear that our approach is the preferred choice. 

Our metric is also preferable, as it is less sensitive to domain. To numerically demonstrate this, we measure the domain sensitivity of the evaluated metrics as the ratio of best Spearman's correlation to worst Spearman's correlation -- this value should be positive (i.e., there is no dataset where the metric becomes negatively correlated), and as close to $1$ as possible (i.e., there is no difference in performance). Looking at Table \ref{tab:all_valid}, we find IDK strongly outperforms all prior metrics, reducing this ratio by more than $37\%$-$66\%$ compared to the best baseline.

\begin{table}[ht]
	\scriptsize
		\centering
		\begin{tabular}{|l|r|}
			\hline
			\textbf{Prior Metric} & Ratio\\
			\hline 
			FED-CORRECT     &  $-0.7$\\
			FED-RELEVANT    &  $-0.7$\\ 
			NORM-PROB       &  $-2.7$\\ 
			COS-NSP-BERT    &  $-7.5$\\
			COS-FT          &  $-13$\\
			GRADE           &  $\infty$\\
			DYNA-EVAL       &  $\infty$\\
			NUP-BERT (TC-S) &  $11.6$\\
			NUP-BERT (TC)   &  $10.3$\\
			COS-MAX-BERT    &  $6.7$\\
			NUP-BERT (H)    &  $6.2$\\
			\hline 
			IDK (H)    &  \textbf{3.9}\\
		    IDK (TC-S) &  \textbf{3.9}\\
			\hline 
		\end{tabular}
	\caption{Ratio of best Spearman correlation to worst on all datasets for all metrics. Sorted in improving order.}
	\label{tab:best-to-worst}
\end{table}

\subsection{Testing NSP feature dimensionality}

As a followup experiment, we tested our assumption that only a fraction of the BERT-NSP features are needed. Plotting the weights learned by IDK on HUMOD, we found a skewed distribution with a small fraction of weights with magnitude above $0.01$ (See Appendix, Figure \ref{fig:nsp_feat_hist}). Hypothesizing that the largest weights correspond to the relevant dimensions, we modified the pretrained huggingface NSP BERT to zero all dimensions of the NSP feature, except for the $7$ dimensions corresponding to the largest IDK HUMOD weights. We then evaluated NSP accuracy on three NLTK \cite{nltk} corpora: Brown, Gutenburg, and Webtext. As expected, we found that reducing the dimensionality from $768$ to $7$ had no negative impact (see Appendix, Table \ref{tab:nsp_feat}). Again, note that the mask was created using IDK trained on HUMOD data, and the weights of BERT and the NSP prediction head were in no way changed. Therefore, it is clear that (at least on these datasets) over $99\%$ of the BERT NSP feature dimensions can be safely discarded.

\subsection{Ablation tests}

\begin{table*}[ht]
	\scriptsize
	\begin{adjustwidth}{-2cm}{-2cm}
		\centering
		\begin{tabular}{|l|c|c|rr|rr|rr|rr|rr|}
			\hline
			\multicolumn{3}{|c|}{}& \multicolumn{2}{|c|}{HUMOD} & \multicolumn{2}{|c|}{USR-TC} & \multicolumn{2}{|c|}{P-DD} & \multicolumn{2}{|c|}{FED-Correctness} & \multicolumn{2}{|c|}{FED-Relevance} \\
			\hline
			\textbf{Data} & L1 & idk & S  & P  & S & P & S & P & S & P & S & P\\
			\hline 
			H & \checkmark & \checkmark & *0.58 (0.00)  & *0.58 (0.00)   & 0.18 (0.00)  & *0.24 (0.00)  & *0.53 (0.00)  & *0.48 (0.01)  & *0.15 (0.00)  & *0.23 (0.00)  & *0.24 (0.00)  & *0.29 (0.00)  \\
			H &  & \checkmark & *0.42 (0.06)  & *0.42 (0.05)   & *0.24 (0.00) & *0.25 (0.00)  & *0.29 (0.06)  & *0.32 (0.03)  & *0.14 (0.00)  & *0.17 (0.01)  & *0.21 (0.01)  & *0.19 (0.02)  \\
			H & \checkmark & & *0.61 (0.00)  & *0.61 (0.00)  & 0.12 (0.00)  & *0.21 (0.01)  & *0.55 (0.00)  & *0.52 (0.01)  & 0.09 (0.00)   & *0.19 (0.01)  & *0.17 (0.00)  & *0.26 (0.01)  \\
			H & &  & *0.60 (0.00)  & *0.61 (0.00)   & 0.18 (0.00)  & *0.26 (0.01)  & *0.54 (0.00)  & *0.50 (0.01)  & 0.10 (0.02)   & \Ankh 0.11 (0.02)  & \Ankh 0.14 (0.02)  & 0.09 (0.03)   \\
			\hline 
			TC-S & \checkmark & \checkmark & *0.58 (0.00)  & *0.58 (0.00)   & 0.18 (0.00)  & *0.22 (0.00)  & *0.54 (0.01)  & *0.49 (0.01)  & *0.15 (0.00)  & *0.23 (0.00)  & *0.24 (0.00)  & *0.29 (0.00)  \\
			TC-S &  & \checkmark & *0.36 (0.04)  & *0.34 (0.05)   & 0.17 (0.01)  & 0.11 (0.01)   & *0.34 (0.03)  & *0.32 (0.04)  & *0.14 (0.00)  & *0.15 (0.01)  & *0.21 (0.00)  & *0.17 (0.01)  \\
			
			TC-S & \checkmark & & *0.59 (0.01)  & *0.54 (0.03)  & \Ankh 0.18 (0.04) & *0.27 (0.02)  & *0.52 (0.03)  & *0.43 (0.05)  & \Ankh 0.14 (0.01)  & *0.21 (0.00)  & *0.22 (0.01)  & *0.29 (0.01)  \\
			TC-S & &   & *0.35 (0.07)  & *0.41 (0.01)   & \Ankh 0.13 (0.10) & *0.21 (0.03)  & \Ankh 0.23 (0.10)  & \Ankh 0.27 (0.11)  & 0.05 (0.06)   & 0.11 (0.03)   & \Ankh 0.12 (0.12)  & \Ankh 0.18 (0.04)  \\
			\hline 
		\end{tabular}
	\end{adjustwidth}
	\caption{Test correlation of various ablations of the proposed metric. The L1 column signifies whether L1 regularization is used ($\lambda = 1$), and the \enquote{idk} column indicates whether the negative samples are \enquote{i don't know}, or a random shuffle of $3,750$ other human responses. Note that L1 regularization is beneficial when training on TC-S. 
	}
	\label{tab:idk_ablate}
\end{table*}

Table \ref{tab:idk_ablate} outlines correlation when ablating the L1 regularization, or when using randomly sampled negative samples in place of ``i don't know". %Specifically, we produce negative examples 
Random samples are produced by shuffling the responses of the next $3,750$ dialogues in the dataset. 

Overall, it appears that the majority of the performance gains come from the combination of L1 regularization with pretrained BERT NSP features. %Going into more detail, t
The clearest observation is that L1 regularization is critical to good performance when using ``i don't know" in place of %negative 
random samples -- otherwise, the model presumably overfits. Second, using ``i don't know" in place of %negative
random samples has a mixed, but relatively minor effect. Thirdly, the effect of L1 regularization is quite positive when training on TC data (regardless of the negative samples), and mixed but smaller when training on HUMOD data. Overall, this suggests that when a validation set of domain-specific annotated relevance data is not available, then L1 regularization may be helpful. Its effect varies by domain, but appears to have a much stronger positive effect than a negative effect. 

The result that L1 regularization allows us to use \enquote{i don't know} in place of random negatives samples is quite interesting, as it seems to counter work in contrastive representation learning \cite{robinson2021contrastive}, and dialogue quality evaluation \cite{pone} suggesting that \enquote{harder} negative examples are better. We believe that the reason for this apparent discrepancy is that \textit{we are not performing feature learning}; the feature space is fixed, pretrained, BERT NSP. Furthermore, we've shown %that the pretrained NSP 
that this feature space is effectively 7 dimensional% (possibly less)
. As a result, we believe that the L1 regularization causes an effective projection to 7D. Consequently, as our model is low-capacity, \enquote{i don't know} is sufficient to find the separating hyperplane. Having said this, it is still unclear why we see \textit{improved} performance on FED when training on HUMOD data. Comparing the histograms of learned weight magnitudes (see Appendix, Figure \ref{fig:nsp_feat_hist_ablate}) we find that the ablated model has larger number of large weights -- we speculate that the random negative samples' variation in irrelevant aspects such as syntactic structure is responsible.

\subsection{Additional Experiments}

We repeated our IDK experiments with two different fixed negative samples; performance and domain sensitivity are generally comparable, although unexpectedly more sensitive to the choice of training data (see Appendix \ref{sec:app_neg}). We also experimented with using the pretrained BERT NSP predictor as a measure of relevance, however performance is considerably worse on the longer-context FED dataset (see Appendix \ref{sec:app_nsp}). Finally, we observed that BCE loss encourages the model to always map \enquote{i don't know} to zero; yet, the relevance of \enquote{i don't know} varies by context. Unfortunately, experiments with a modified triplet loss did not yield improvements (see Appendix \ref{sec:app_triplet}).

\section{Related Work}

In addition to the prior metrics already discussed, the area of dialogue relevance is both motivated by, and jointly developed with, the problem of automatic dialogue evaluation. As relevance is a major component of good dialogue, there is a bidirectional flow of innovations. The NUP-BERT relevance metric is very similar to BERT-RUBER \cite{ghazarian-etal-2019-better}; both train a small MLP to perform the next-utterance-prediction task based on aggregated BERT features. Both of these share a heritage with earlier self-supervised methods, such as adversarial approaches to dialogue evaluation that train a classifier to distinguish human from generated samples \cite{adversarial}. Another example of shared development is the use of word-overlap metrics such as BLEU \cite{papineni-etal-2002-bleu} and ROUGE \cite{lin-2004-rouge} that have been imported wholesale into both dialogue relevance and overall quality from the fields of machine-translation and summarization, respectively. 

Simultaneously, metrics of dialogue evaluation have been motivated by dialogue relevance. There is a long history of evaluating dialogue models on specific aspects; \citet{finch-choi-2020-towards} performed a meta-analysis of prior work, and proposed  dimensions of: grammaticality, relevance, informativeness, emotional understanding, engagingness, consistency, proactivity, and satisfaction. New approaches to dialogue evaluation have emerged from this body of work, seeking to aggregate individual measures of various dimensions of dialogue, often including relevance \cite{mehri-eskenazi-2020-usr, phy-etal-2020-deconstruct, ling_feat}. These approaches also share heritage with earlier ensemble measures of dialogue evaluation such as RUBER \cite{ruber} -- although in the case of RUBER, it combined a referenced and unreferenced metric rather than separate aspects.

Metrics of dialogue relevance and quality also share common problems such as the diversity of valid responses. Our findings that existing relevance metrics generalize poorly to new domains is consistent with previous findings about metrics of dialogue quality  \cite{adem_retro, yeh-etal-2021-comprehensive}.  Thus, our work suggests that this challenge extends to the subproblem of dialogue relevance as well. 

%Interestingly, \citet{yeh-etal-2021-comprehensive} found that trained quality metrics tended to perform better on the datasets they were trained on. %we do not find this trend for relevance when comparing NUP-BERT or IDK on HUMOD or USR-TC. Futhermore, it is interesting to note that 
%For NUP-BERT, we did not find this effect on HUMOD, although we did find variations in performance based on training domain. However, IDK appears to eliminate these variations.

At the same time, it must be remembered that measuring holistic dialogue quality is a very different task from measuring dialogue relevance -- it is well established that aspects of dialogue such as fluency, and interestingness are major components of quality \cite{mehri-eskenazi-2020-usr, mehri-eskenazi-2020-unsupervised}, and these should have no impact on relevance.

With respect to prior work comparing relevance metrics, we are aware of only one tangential work. \citet{yeh-etal-2021-comprehensive} performed a comparison of various metrics of dialogue \textit{quality}; within this work they dedicated three paragraphs to a brief comparison of how these \textit{quality} metrics performed at predicting various dialogue qualities, including relevance. They reported results on only two of the datasets we used (P-DD and FED). Interestingly, the authors found that the FED metric performs well on P-DD (reporting a Spearman's correlation of 0.507), however our results demonstrate that the \textit{components} of FED that are meant to measure relevance (i.e. FED-REL and FED-COR) are significantly \textit{negatively} correlated with human relevance scores. Additionally, as \citet{yeh-etal-2021-comprehensive} focus on quality, they do not compare performance between the two relevance datasets. Instead they compare performance on quality against performance on relevance, and use the discrepancy to conclude that measuring relevance alone (as done by NORM-PROB) is insufficient to determine quality. Although we agree that relevance alone is insufficient for dialogue quality evaluation, our work provides a richer understanding. Our finding that NORM-PROB performs poorly across a range of relevance datasets suggests that the poor performance of NORM-PROB in the quality-prediction task is also caused by the \textit{poor relevance generalization} in addition to the insufficiency of relevance to measure overall quality. %Finally, whereas \citet{yeh-etal-2021-comprehensive} used model checkpoints on different datasets, we train all of the trainable models studied on HUMOD, USR-TC, and a smaller subset of USR-TC, giving a better idea of how training data affects performance.

%TODO TODO TODO: try averaging all out and see what's what w.r.t. alleged perf on P-DD? ANSWER: I think they used semanticaly appropriate; also they didn't put model into eval mode so there's some noise

%Also, unlike ourselves, they do not study the effects of training the same metric on different datasets. 

% TODO: BS: reduced effect of training set of performance. (therefore reduced effect of train set, if not test set -- PROGRESS!!!)

% TODO: Update validation perf. table with FED results

\section{Discussion}
Our experiments demonstrate that several published measures of dialogue relevance have poor, or even negative, correlation when evaluated on new datasets of dialogue relevance, suggesting overfitting to either model or domain. As such, it is clear that further research into new measures of dialogue relevance is required, and that %great
care must be taken in their evaluation to compare against a number of different models in a number of domains. Furthermore, it is also clear that for the current practitioner who requires a measure of relevance, there are no guarantees that current methods will perform well on a given domain. As such, it is wise to collect a validation dataset of human-annotated relevance data for use in selecting a relevance metric. If this is not possible, then our metric, IDK, appears to be the best option -- achieving both good correlation and the lowest domain sensitivity, even when trained on different domains. Furthermore, when training data is scarce, our results suggest that the use of strong regularization allows for the use of a single negative example, \enquote{i don't know}, in the place of randomly sampled negative samples. If that is still too data intensive, then our results suggest that our metric is fairly agnostic to the domain of the training data; therefore training data can be used from a different dialogue domain in place of the domain of interest. 

Having said this, it is clear that further research into what exactly these metrics are measuring, and why they fail to generalize, is merited. The results are often counter-intuitive; our demonstration that $99\%$ of the BERT NSP features can be safely discarded is just one striking example. Similarly, although our empirical results suggest that use of a single negative example generalizes across domains, there is no compelling theoretical reason why this should be so. More generally, all the metrics outlined are complex, dependent on large corpora, and created without ground truth annotations. As a result, they are all dependent on either surrogate tasks (i.e., NUP), or unsupervised learning (e.g., FastText embeddings). Consequently, it is especially difficult to conclude what exactly these metrics are measuring. At present, the only strong justification that these metrics are indeed measuring relevance is good correlation with human judgements -- poor generalization across similar domains is not an encouraging result.  

Although the metric outlined is not appropriate for final model evaluation (as it risks unfairly favouring dialogue models based on the same pretrained BERT, or similar architectures), our aim is to provide a useful metric for  rapid prototyping and hyperparameter search. Additionally, we hope that our findings on the domain sensitivity of existing metrics will spur further research into both the cause of -- and solutions to -- this problem.

\section{Conclusion}

Our work demonstrates that several existing metrics of dialogue relevance are problematic as their performance varies wildly between test-domains. We take a first step towards resolving this issue by proposing IDK: a simple metric that is less sensitive to test domain and trainable with minimal data. We reduce IDK's data requirements through the novel use of a fixed negative example, provide evidence that the underlying BERT NSP features are low-dimensional, and propose that this fact (combined with IDK's lack of feature learning) allows for the counter-intuitive use of a single negative example. Beyond this, we call for better evaluation of future relevance metrics, and thus release our code for processing four diverse,  publicly available, relevance-annotated data sets.

\section{Acknowledgements}

Resources used in preparing this research were provided, in part, by the Province of Ontario, the Government of Canada through CIFAR, and companies sponsoring the Vector Institute \url{https://vectorinstitute.ai/partners/}. Ian Berlot-Attwell is funded by an Ontario Graduate Scholarship and a Vector Institute Research Grant. Frank Rudzicz is supported by a CIFAR Chair in AI. We would also like to thank the various reviewers who helped to shape and improve this work; without them it would not be what it is today.

%TODO: First page -- example of failures/successes?  Add scatter plot comparing human scores against a couple of the metrics?

%TODO: If time, paper proposing cosine-similarity on fine-tuned BERT embeddings https://link-springer-com.myaccess.library.utoronto.ca/chapter/10.1007%2F978-981-15-8395-7_5

\bibliography{anthology,custom}
\bibliographystyle{acl_natbib}

\clearpage
\pagenumbering{arabic} 
\appendix

\section{Example Evaluations}\label{app:example_eval}

\begin{table}[ht]
	\centering
	\begin{tabular}{|p{4.75cm}|r|r|}
		\hline 
		Dialogue Turn  & Human & IDK \\
		\hline 
		\hline 
		Mommy -- & - & - \\
		Yes, darling.  & - & - \\
		Did you ever make a wish?  & - & - \\
		Oh, lots of times.  & - & - \\
		\hline 
		Did your wishes ever come true? & 5.00 & 4.97 \\
		\textit{What's your real name?} & 1.00 & 3.81 \\
		\hline 
		Sometimes. & 4.67 & 4.60 \\
		\textit{From high school Mary? Yeah, I saw her about six months ago at a convention in Las Vegas.} & 1.00 & 1.13 \\
		\hline 
		I made a wish today, and it came true just like Edward said it would. & 5 & 4.9 \\
		\textit{When I am sure I am among friends.} & 2.33 & 3.01 \\ 
		\hline 
		\hline 
		Yes, Albert? & - & - \\
		John, we're going huntin'. & - & - \\
		Who's goin? & - & - \\
		We're all going. & - & - \\
		\hline 
		Nick's going? & 4.67 & 4.65 \\
		\textit{I will keep you safe. We are both older.} & 2.00 & 1.09 \\
		\hline 
		Nick , Vince , Albert and John. & 4.00 & 4.95 \\
		\textit{A ride? Hell, that's a good idea. Okay, let's go. Hey, let's go.} & 2.33 & 4.68 \\
		\hline 
		No women? & 4.00 & 2.39 \\
		\textit{I guess so} & 3.00 & 2.59 \\
		\hline 
	\end{tabular}
	\caption{Two multi-turn examples from HUMOD test set. The randomly sampled distractor turns are italicized, and are not part of the context in subsequent turns. For ease of comparison, the  scores generated by our metric (IDK trained on HUMOD) are linearly shifted and re-scaled to 1-5.}
	\label{tab:idk_example}
\end{table}

\section{NSP Masking Experiment Results}

\begin{table}[h!]
	\centering
	\begin{tabular}{|c|r|r|r|}
		\hline
		Masked & Brown & Gutenburg & Webtext \\
		\hline
		& \textbf{85.7\%} & 75.3\% &  65.4\% \\
		\checkmark & 85.6\% & \textbf{75.5\%} & \textbf{68.2\%} \\
		\hline 
	\end{tabular}
	\caption{Next Sentence Prediction (NSP) performance on various NLTK \cite{nltk} corpora using a pre-trained BERT and NSP head. When masked, we zero-out the 768-dim BERT NSP feature, leaving only the 7 dimensions corresponding to the largest magnitude weights in IDK (H) (i.e., we zero out $>99\%$ of the feature vector).}
	\label{tab:nsp_feat}
\end{table}

The results of the NSP masking experiment are outlined in Table \ref{tab:nsp_feat}. Note that masking $>99\%$ of the NSP feature had no impact on the pretrained model, and actually improved accuracy by $2.8\%$ on the Webtext corpus.

\section{Exact objectives of prior metrics}\label{app:purpose}

In this section, we briefly outline the stated purpose of each of our relevance metrics evaluated:

\begin{itemize}
	\item COS-FT: \enquote{In this work, given a dialogue history, we regard as a coherent response an utterance that is thematically correlated and naturally continuing from the previous turns, as well as lexically diverse.} \cite{xu-etal-2018-better}
	\item NUP-BERT: \enquote{Maintains Context: Does the response serve as a valid continuation of the preceding conversation?} \cite{mehri-eskenazi-2020-usr}
	\item NORM-PROB: \enquote{context coherence of a dialogue: the meaningfulness of a response within the context of prior query} \cite{pang-etal-2020-towards}
	\item FED-REL: \enquote{Is the response relevant to the conversation?} \cite{mehri-eskenazi-2020-unsupervised}
	\item FED-COR: \enquote{Is the response correct or was there a misunderstanding of the conversation? [...] No one has specifically used Correct, however its meaning is often encapsulated in Relevant.} \cite{mehri-eskenazi-2020-unsupervised}
\end{itemize}

We also outline the stated purpose of the \textit{dialogue coherence} metrics evaluated:

\begin{itemize}
	\item GRADE: \enquote{Coherence, what makes dialogue utterances unified rather than a random group of sentences} \cite{huang-etal-2020-grade}
	\item DYNA-EVAL: \enquote{dialogue coherence: considers whether a piece of text is in a consistent and logical manner, as opposed to a random collection of sentences} \cite{zhang-etal-2021-dynaeval} 
\end{itemize}

% Xu: Coherence: "In this work, given a dialogue history, we regard as a coherent response an utterance that is thematically correlated and naturally continuing from the previous turns, as well as lexically diverse."

% USR: Maintains Context: Does the response serve as a valid continuation of the preceding conversation?

% Pang: context coherence of a dialogue: the meaningfulness of a response within the context of prior query

% FED: \say{Is the response relevant to the conversation?} and \say{Is the response correct or was there a misunderstanding of the conversation? }

% GRADE: dialogue Coherence, what makes dialogue utterances unified rather than a random group of sentences

% DynaEval: dialogue coherence: considers whether a piece of text is in a consistent and logical manner, as opposed to a random collection of sentences.

\section{Details for Prior work}

\subsection{NORM-PROB}\label{app:pang}

\citet{pang-etal-2020-towards} relied on a pretrained generative language model (specifically GPT-2). Their proposed metric is the conditional log-probability of the response given the context, normalized to the range $[0,1]$. Specifically, for a context $q$ with candidate response $r$, their proposed relevance score is defined as: $c(q\,|\,r) = - \frac{\max(c_{5th}, \frac{1}{|r|} \log P(r\,|\,q)) - c_{5th}}{c_{5th}}$, where $|r|$ is the number of tokens in the response, $P(r\,|\,q)$ is the conditional probability of the response given the context under the language model, and $c_{5th}$ is the $5^{th}$ percentile of the distribution of $\frac{1}{|r|} \log P(r\,|\,q)$ over the examples being evaluated.

\section{Learned HUMOD-IDK Weights}

\begin{figure}[h!]
	\centering
	\includegraphics[width=1\linewidth]{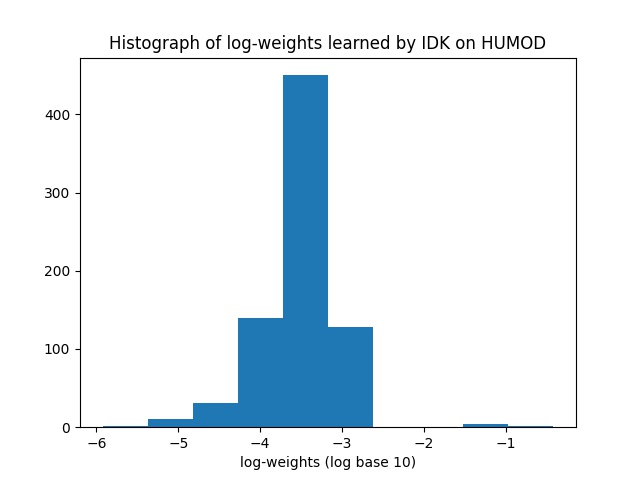}
	\caption{Histogram of log weight magnitudes learned by IDK on HUMOD. Note the small number of weights that are an order of magnitude larger.}
	\label{fig:nsp_feat_hist}
\end{figure}

\begin{figure}[h!]
	\centering
	\includegraphics[width=1\linewidth]{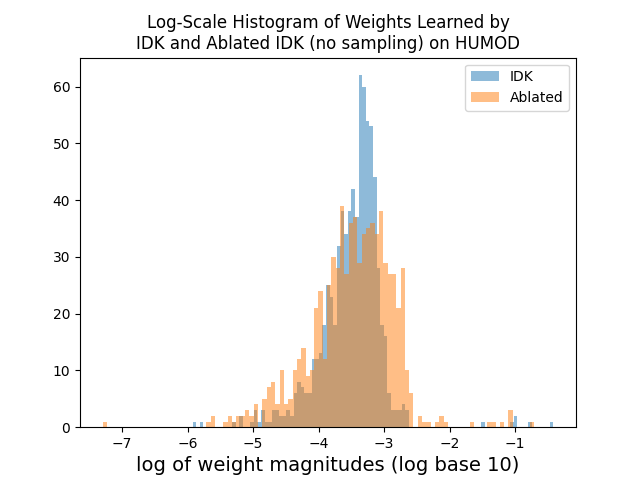}
	\caption{Histogram of log weight magnitudes learned by IDK and Ablated IDK on HUMOD. The specific ablation is the use of random negative samples in place of \enquote{i don't know}. Note that Ablated IDK has a larger number of large weights than normal IDK.}
	\label{fig:nsp_feat_hist_ablate}
\end{figure}

Figure \ref{fig:nsp_feat_hist} depicts the distribution of weight-magnitudes learned by IDK on the HUMOD training set. Notably, there is a very small subset of weights which is an order of magnitude larger than the others. Figure \ref{fig:nsp_feat_hist_ablate} demonstrates that the use of random sampling in place of \enquote{i don't know} when training on the HUMOD dataset causes a larger number of large weights.

\section{Scatter Plots}\label{app_scatter}

\begin{figure}[h!]
    \centering
    \includegraphics[width=\linewidth]{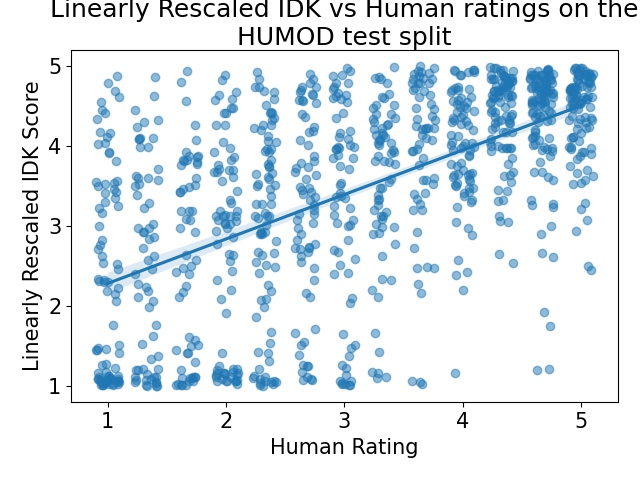}
    \caption{IDK scores, linearly re-scaled to the range 1-5, versus human scores of relevance, on the HUMOD test set.}
    \label{fig:humod}
\end{figure}

\begin{figure}[h!]
    \centering
    \includegraphics[width=\linewidth]{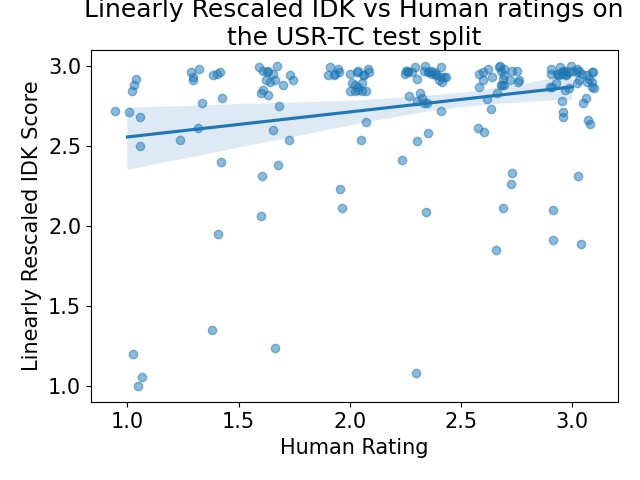}
    \caption{IDK scores, linearly re-scaled to the range 1-3, versus human scores of relevance, on the USR-TC test set.}
    \label{fig:usr}
\end{figure}

\begin{figure}[h!]
    \centering
    \includegraphics[width=\linewidth]{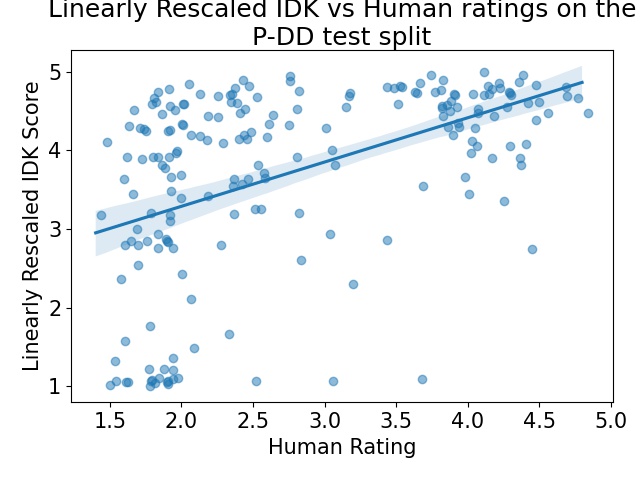}
    \caption{IDK scores, linearly re-scaled to the range 1-5, versus human scores of relevance, on the P-DD test set.}
    \label{fig:pdd}
\end{figure}

\begin{figure}[h!]
	\centering
	\includegraphics[width=\linewidth]{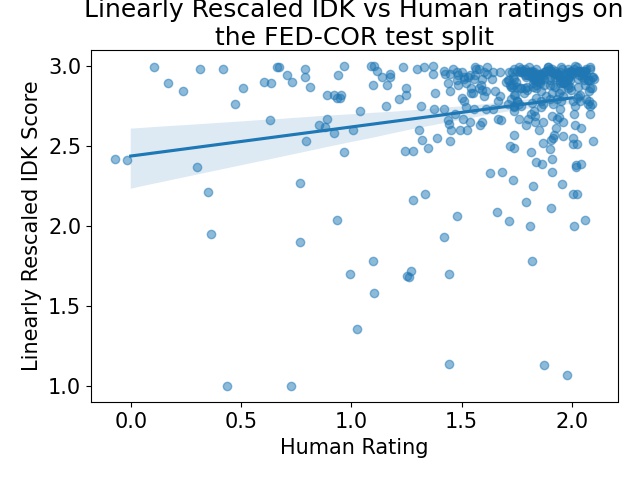}
	\caption{IDK scores, linearly re-scaled to the range 1-3, versus human scores of relevance, on the FED-CORRECT test set.}
	\label{fig:fed_cor}
\end{figure}

\begin{figure}[h!]
	\centering
	\includegraphics[width=\linewidth]{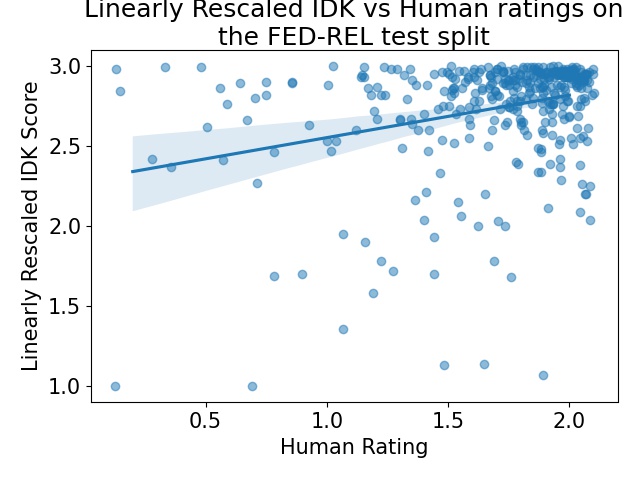}
	\caption{IDK scores, linearly re-scaled to the range 1-3, versus human scores of relevance, on the FED-RELEVANT test set.}
	\label{fig:fed_rel}
\end{figure}

Figures \ref{fig:humod}, \ref{fig:usr}, \ref{fig:pdd}, \ref{fig:fed_cor}, and \ref{fig:fed_rel} illustrate IDK vs human scores of relevance, where the IDK training data is HUMOD. A regression line is fitted to highlight the trend.

\FloatBarrier

\section{Performance on validation data split}
\label{sec:appendix}

Correlations of the models on the validation set are outlined in Table \ref{tab:valid_prior} for prior metrics, and in Table \ref{tab:all_valid} for all ablations and variants of our model.

\begin{table*}%[ht]
\centering
\begin{tabular}{|l|rr|rr|}
\hline
 & \multicolumn{2}{|c|}{HUMOD} & \multicolumn{2}{|c|}{USR-TC}\\
\hline
\textbf{Prior Metric} & S & P  & S & P  \\
\hline 
COS-FT      & $0.08$ & $0.08$ & *$0.27$ & $0.17$ \\
COS-MAX-BERT    & $0.08$ & $0.05$  & $0.18$ & *$0.19$ \\
COS-NSP-BERT    & $0.06$ & *$0.09$  &  *$0.23$ & *$0.25$ \\
NORM-PROB   & *$0.27$ & *$0.25$ & *$ -0.29$ & *$ -0.30$  \\ 
FED-CORRECT & *$-0.10$ & *$-0.09$ & $-0.14$ & $-0.15$ \\
FED-RELEVANT& *$-0.10$ & *$-0.09$ & $-0.14$ & $-0.16$ \\
\hline 
GRADE & *0.64      & *0.64      & 0.02        & 0.00        \\
DYNA-EVAL &*0.14      & *0.15      & $-0.05$       & $-0.06$       \\
\hline
NUP-BERT (H)    & *0.37 (0.01) & *0.38 (0.00) & *0.38 (0.02) & *0.39 (0.01) \\
NUP-BERT (TC-S)   & *0.32 (0.01) & *0.36 (0.02) & *0.38 (0.04) & *0.41 (0.04) \\
NUP-BERT (TC)    & *0.33 (0.02) & *0.37 (0.02) & *0.45 (0.07) & *0.44 (0.02) \\
\hline 
\end{tabular}
\caption{Spearman (S) and Pearson (P) correlations of prior metrics with human ratings on the validation splits of all provided dataset. As NUP-BERT is trained we perform 3 runs, reporing the mean and standard deviation. (*) denotes $p< 0.01$ accross all trials. Underline indicates a negative correlation. NOTE: USR scores are human only for COS-FT, NORM-PROB and NUP-BERT}
\label{tab:valid_prior}
\end{table*}
  
  % NOTE: original message: NOTE: USR scores are human only for all except COS-BERT’s
  
\begin{table*}%[hb!]
	\scriptsize
	\begin{adjustwidth}{-2cm}{-2cm}
		\centering
		\begin{tabular}{|l|c|c|rr|rr|rr|rr|rr|}
			\hline
			\multicolumn{3}{|c|}{}& \multicolumn{2}{|c|}{HUMOD} & \multicolumn{2}{|c|}{USR-TC} & \multicolumn{2}{|c|}{P-DD} & \multicolumn{2}{|c|}{FED-Correctness} & \multicolumn{2}{|c|}{FED-Relevance} \\
			\hline
			\textbf{Data} & L1 & idk & S  & P  & S & P & S & P & S & P & S & P\\
			\hline 
			H  & \checkmark & \checkmark & *0.59 (0.01)  & *0.55 (0.02)  & 0.17 (0.01)  & *0.28 (0.01)  & *0.54 (0.03)  & *0.44 (0.02)  & \Ankh 0.13 (0.02)  & *0.21 (0.01)  & *0.21 (0.01)  & *0.30 (0.00)  \\
			H  &  & \checkmark & *0.15 (0.05)  & *0.19 (0.06)  & \Ankh 0.19 (0.01) & *0.25 (0.02)  & 0.10 (0.04)   & \Ankh 0.17 (0.05)  & 0.10 (0.02)   & \Ankh 0.11 (0.02)  & \Ankh 0.14 (0.02)  & 0.09 (0.03)   \\
			H  & \checkmark & & *0.45 (0.24)  & *0.42 (0.21)   & 0.14 (0.04)  & \Ankh 0.23 (0.10)  & \Ankh 0.39 (0.21)  & *0.34 (0.14)  & 0.11 (0.02)   & \Ankh 0.18 (0.06)  & *0.20 (0.02)  & *0.25 (0.08)  \\
			H  & &  & *0.61 (0.00)  & *0.60 (0.01)  & 0.17 (0.00)  & *0.23 (0.01)  & *0.55 (0.01)  & *0.53 (0.01)  & \Ankh 0.14 (0.00)  & *0.20 (0.02)  & *0.22 (0.00)  & *0.27 (0.02)  \\
			\hline 
			TC-S  & \checkmark & \checkmark & *0.32 (0.44)  & *0.25 (0.55)  & 0.12 (0.06)  & \Ankh 0.10 (0.24)  & *0.24 (0.47)  & *0.21 (0.46)  & 0.10 (0.04)   & \Ankh 0.10 (0.14)  & \Ankh 0.17 (0.07)  & \Ankh 0.14 (0.21)  \\
			TC-S  &  & \checkmark & *0.27 (0.11)  & *0.26 (0.10)  & 0.16 (0.02)  & 0.14 (0.03)   & \Ankh 0.22 (0.12)  & \Ankh 0.22 (0.09)  & \Ankh 0.13 (0.01)  & *0.15 (0.01)  & *0.19 (0.02)  & *0.17 (0.02)  \\
			TC-S  & \checkmark & & *-0.20 (0.69) & *-0.20 (0.65) & -0.03 (0.17) & \Ankh -0.05 (0.29) & *-0.18 (0.62) & *-0.19 (0.54) & \Ankh -0.05 (0.18) & *-0.07 (0.26) & *-0.08 (0.27) & *-0.09 (0.35) \\
			TC-S  & &   & \Ankh 0.18 (0.20)  & *0.18 (0.06)  & 0.04 (0.07)  & 0.09 (0.17)   & 0.10 (0.07)   & 0.07 (0.06)   & 0.02 (0.10)   & \Ankh 0.08 (0.07)  & 0.00 (0.10)   & \Ankh 0.12 (0.10)  \\
			\hline 
		\end{tabular}
	\end{adjustwidth}
	\caption{Repeat of ablation experiments, instead using modified triplet loss ($m=0.4$) in place of BCE. Contrary to our intuition, we do not find any improvement in performance. Comparing against Table \ref{tab:idk_ablate}, we find either equivalent or degraded performance, with an additional tendency to converge to a degenerate solution (e.g., see high variances in TC-S with L1 and idk).}
	\label{tab:triplet}
\end{table*}

\begin{table*}
\begin{tabular}{lllll}
	\hline
	Name                 & HUMOD Spear       & HUMOD Pear        & TC Spear          & TC Pear           \\
	H\_Rand3750\_bce       & *0.58 (0.00)      & *0.57 (0.01)      & *0.46 (0.00)      & *0.43 (0.02)      \\
	H\_Rand3750           & *0.58 (0.00)      & *0.58 (0.00)      & *0.46 (0.00)      & *0.45 (0.02)      \\
	H\_IDK\_L1             & *0.56 (0.01)      & *0.53 (0.02)      & *0.45 (0.03)      & *0.44 (0.02)      \\
	H\_IDK\_L2             & *0.55 (0.00)      & *0.55 (0.01)      & *0.44 (0.00)      & *0.44 (0.00)      \\
	H\_Rand3750\_L1        & *0.42 (0.22)      & *0.40 (0.20)      & *0.44 (0.00)      & *0.45 (0.01)      \\
	H\_Rand3750\_L2        & *0.56 (0.00)      & *0.55 (0.01)      & *0.45 (0.00)      & *0.44 (0.02)      \\
	H\_Rand3750\_bce\_L1    & *0.58 (0.00)      & *0.58 (0.00)      & *0.45 (0.00)      & *0.46 (0.00)      \\
	H\_Rand3750\_bce\_L2    & *0.57 (0.00)      & *0.56 (0.00)      & *0.45 (0.00)      & *0.42 (0.00)      \\
	H\_IDK\_bce\_L1         & *0.57 (0.00)      & *0.56 (0.00)      & *0.42 (0.01)      & *0.41 (0.00)      \\
	H\_IDK\_bce\_L2         & *0.50 (0.01)      & *0.51 (0.01)      & *0.39 (0.00)      & *0.42 (0.00)      \\
	H\_IDK\_bce            & *0.39 (0.05)      & *0.40 (0.05)      & *0.36 (0.02)      & *0.34 (0.00)      \\
	H\_IDK                & *0.15 (0.05)      & *0.19 (0.06)      & 0.09 (0.05)       & \Ankh 0.21 (0.05)  \\
	TC-S\_IDK\_L1          & *0.29 (0.43)      & *0.23 (0.53)      & *0.39 (0.07)      & *0.41 (0.07)      \\
	TC-S\_IDK\_L2          & *0.54 (0.01)      & *0.55 (0.01)      & *0.43 (0.01)      & *0.44 (0.00)      \\
	TC-S\_IDK\_bce\_L1      & *0.57 (0.00)      & *0.56 (0.00)      & *0.43 (0.00)      & *0.40 (0.00)      \\
	TC-S\_IDK\_bce\_L2      & *0.47 (0.02)      & *0.48 (0.01)      & *0.41 (0.00)      & *0.39 (0.01)      \\
	TC-S\_IDK\_bce         & *0.35 (0.04)      & *0.33 (0.05)      & *0.40 (0.01)      & *0.31 (0.01)      \\
	TC-S\_IDK             & *0.25 (0.10)      & *0.24 (0.10)      & *0.34 (0.05)      & *0.36 (0.03)      \\
	TC-S\_Rand3750\_L1     & *-0.19 (0.67)     & *-0.20 (0.63)     & *-0.13 (0.52)     & *-0.14 (0.50)     \\
	TC-S\_Rand3750\_L2     & \Ankh -0.33 (0.27) & \Ankh -0.32 (0.26) & *-0.45 (0.02)     & *-0.43 (0.02)     \\
	TC-S\_Rand3750\_bce\_L1 & *0.56 (0.01)      & *0.52 (0.03)      & *0.44 (0.03)      & *0.40 (0.02)      \\
	TC-S\_Rand3750\_bce\_L2 & *0.04 (0.55)      & *0.09 (0.56)      & \Ankh -0.26 (0.27) & \Ankh -0.23 (0.31) \\
	TC-S\_Rand3750\_bce    & *0.31 (0.05)      & *0.36 (0.03)      & \Ankh 0.16 (0.29)  & \Ankh 0.18 (0.26)  \\
	TC-S\_Rand3750        & \Ankh 0.15 (0.17)  & *0.11 (0.02)      & \Ankh -0.14 (0.24) & \Ankh -0.06 (0.27) \\
	\hline
\end{tabular}

\caption{Validation correlation of all of tested variants and ablations of our model. H vs. TC-S indicates training set (HUMOD or subset of TopicalChat respectively). IDK vs. Rand3750 indicates whether negative examples are \enquote{i don't know} or random. If bce is present, then BCE was used as the loss, otherwise our modified triplet loss is used. If L1 or L2 is present, then L1 or L2 regularization with $\lambda=1$ is used respectively, otherwise no regularization is used. Again, standard deviation over three trials is reported in parentheses, and `*' is used to indicate that all trials were significant at $p< 0.01$. `\Ankh' indicates at least one trial was significantly different from zero at $p<0.01$. Note that L1 and L2 regularization have similar effects, with the exception of worse performance between TC-S\_Rand2750\_bce\_L1 and TC-S\_Rand2750\_bce\_L2; we suspect this could be overcome with hyperparameter tuning.}
\label{tab:all_valid}
\end{table*}

\FloatBarrier

\begin{table*}[ht!]
%\small
\scriptsize
\begin{adjustwidth}{-2cm}{-2cm}
    \centering
    \begin{tabular}{|l|rr|rr|rr|rr|rr|}
\hline
 & \multicolumn{2}{|c|}{HUMOD} & \multicolumn{2}{|c|}{USR-TC} & \multicolumn{2}{|c|}{P-DD} & \multicolumn{2}{|c|}{FED-Correctness} & \multicolumn{2}{|c|}{FED-Relevance}\\
\hline
\textbf{Prior Metric} & S  & P  & S & P & S & P & S & P & S & P\\
\hline 
NUP-BERT (H)    & *0.33 (0.02) & *0.37 (0.02)  & 0.10 (0.02)  & *0.22 (0.01)  & *\textbf{0.62} (0.04) & *0.54 (0.02) & \Ankh 0.14 (0.04) & *0.21 (0.03) & *0.22 (0.01) & *\textbf{0.30} (0.01) \\
NUP-BERT (TC-S) & *0.29 (0.02) & *0.35 (0.03) & \Ankh 0.17 (0.03) & \Ankh 0.20 (0.04)  & *0.58 (0.05) & *0.56 (0.04)  & 0.05 (0.04)  & 0.12 (0.01)  & \Ankh 0.16 (0.04) & *0.21 (0.01) \\
NUP-BERT (TC)   & *0.30 (0.01) & *0.38 (0.00)   & 0.16 (0.02)  & *0.21 (0.02)& *\textbf{0.62} (0.05) & *\textbf{0.58} (0.04) & 0.06 (0.01)  & \Ankh 0.12 (0.02) & *0.18 (0.02) & *0.23 (0.01) \\
\hline
 IDK (H)          & *0.58 (0.00)  & *0.58 (0.00)     & 0.18 (0.00)  & *0.24 (0.00)  & *0.53 (0.00)  & *0.48 (0.01)  & *0.15 (0.00) & *\textbf{0.23} (0.00)  & *\textbf{0.24} (0.00) & *0.29 (0.00) \\
 IDK (TC-S)      & *0.58 (0.00)  & *0.58 (0.00)     & 0.18 (0.00)  & *0.22 (0.00)  & *0.54 (0.01)  & *0.49 (0.01)  & *0.15 (0.00) & *\textbf{0.23} (0.00)  & *\textbf{0.24} (0.00) & *0.29 (0.00) \\
\hline 
IDK-ICS (H) & *0.55 (0.01)  & *0.53 (0.00)  & *\textbf{0.25} (0.01) & *\textbf{0.27} (0.00)  & *0.44 (0.01)  & *0.39 (0.00)  & *\textbf{0.16} (0.00)  & *0.22 (0.00)  & *0.22 (0.00)  & *\textbf{0.30 }(0.00)  \\
IDK-ICS (TC-S) & *0.58 (0.00)  & *0.47 (0.00)  & 0.17 (0.00)  & *\textbf{0.27 }(0.00)  & *0.52 (0.00)  & *0.36 (0.00)  & *0.14 (0.00)  & *0.16 (0.00)  & *0.22 (0.00)  & *0.24 (0.00)  \\
IDK-OK (H) & *0.58 (0.00)  & *\textbf{0.59} (0.00)  & 0.15 (0.00)  & *0.23 (0.00)  & *0.49 (0.00)  & *0.47 (0.00)   & 0.11 (0.00)   & *0.19 (0.00)  & *0.19 (0.00)  & *0.26 (0.00)  \\
IDK-OK (TC-S) & *\textbf{0.59} (0.00)  & *\textbf{0.59} (0.00)  & 0.18 (0.00)  & *0.24 (0.00)  & *0.52 (0.00)  & *0.46 (0.00)  & *0.15 (0.00)  & *\textbf{0.23} (0.00)  & *0.23 (0.00)  & *0.29 (0.00)  \\
\hline
BERT NSP & *\textbf{0.59} & *0.40 & 0.17 & *0.25 & *0.53 & *0.31 & 0.12 & 0.10 & *0.21 & *0.18 \\
\hline 
\end{tabular}
\end{adjustwidth}
    \caption{Comparison of our proposed metric (IDK) against the pretrained BERT NSP predictor on the test set. We also trained IDK with different fixed negative examples, "i couldn't say" (IDK-ICS) and "i'm ok." (IDK-OK). Note BERT NSP tends to have comparable Spearman's performance and worse Pearson's correlation. The only exception is FED where BERT NSP has inferior performance. In general, IDK with different fixed negative samples outperforms NUP-BERT, and is less sensitive to training data, although not to the same extent as baseline IDK.
    }
    \label{tab:app_nsp}
\end{table*}

\FloatBarrier

\section{Additional Experiments: Triplet Loss}\label{sec:app_triplet}

An intuitive limitation of using \enquote{i don't know} as a negative example with BCE loss is that this encourages the model to always map \enquote{i don't know} to exactly zero. However, the relevance of \enquote{i don't know} evidently varies by context. Clearly, it is a far less relevant response to \enquote{I was interrupted all week and couldn't get anything done, it was terrible!} than it is to \enquote{what is the key to artificial general intelligence?} Motivated by this intuition, we experimented with a modified triplet loss, $\mathcal{L} (c,r) = -\log \left(1 + m - f_{t}(c,r) \right)$ where $ f_{\textit{t}}(c,r) =  \max \left( y(c,r) - y(c,r') + m, 0 \right)$.

%\begin{equation}
%	f_{\textit{t}}(c,r) =  \max \left( y(c,r) - y(c,r') + m, 0 \right)
%\end{equation}

%\begin{equation}
%	\mathcal{L} (c,r) = -\log \left(1 + m - f_{t}(c,r) \right)
%\end{equation}

%$$ f_{\textit{t}}(c,r) =  \max \left( y(c,r) - y(c,r') + m, 0 \right)$$
%$$\mathcal{L} (c,r) = -\log \left(1 + m - f_{t}(c,r) \right)$$

Intuitively, a triplet loss would allow for the relevance of \enquote{i don't know} to shift, without impacting the loss as long as the ground-truth responses continue to score sufficiently higher. Note that the loss is modified to combat gradient saturation due to the sigmoid non-linearity. However, the results (see Table \ref{tab:triplet}) suggest equivalence, at best. Often, this loss performs equivalently to BCE but it can also produce degenerate solutions (note the high variance when training on TC data). Furthermore, it does not appear to produce superior correlations. 

For this reason, we believe that, although adapting triplet loss for next-utterance prediction in place of BCE could be made to work, it does not appear to provide any advantages. If validation data is available, it can be used to confirm whether the model has reached a degenerate solution, and thus this loss could be used interchangeably with BCE. However, there does not appear to be any advantage in doing so. 

\FloatBarrier

\section{Additional Experiments: BERT NSP}\label{sec:app_nsp}

As a followup experiment we compared IDK against directly using the pretrained BERT NSP predictor. In general, Spearman's correlation was comparable on all datasets \textit{except for FED}, and Pearson's correlation was degraded. Performance on FED was inferior to IDK. We speculate that the reason for this is that the FED datasets has longer contexts, which is problematic for the NSP predictor as it was trained with sentences rather than utterances. Results are summarized in Table \ref{tab:app_nsp}.

\section{Additional Experiments: IDK with other fixed negative samples}\label{sec:app_neg}

As a followup experiment we trained IDK using two different fixed negative samples: "i couldn't say" (simply chosen as a synonym for "i don't know"), and "i'm ok." (chosen as an example of a generic response from \citet{li-etal-2016-diversity}).
Results are reported in Table \ref{tab:app_nsp}; in general we still see an performance improvement over NUP-BERT, and in some cases we exceed the performance of baseline IDK. We also see that performance remains consistent between runs, maintaining a lower standard deviation than NUP-BERT.

However, it is also clear that changing the fixed negative sample has some unexpected consquences: specifically, we see variation based on training data that is not observed when using "i don't know" as the fixed negative sample (although the variation due to training data appears to be less than NUP-BERT).

We retain the reduced sensitivity to test set. Specically, our ratios of best-to-worst Spearman's correlation are 3.44 for IDK-ICS (H), 4.14 for IDK-ICS (TC-S), 5.27 for IDK-OK (H), and 3.93; most are very close to the baseline IDK ratio of 3.9, and all are an improvement on the best prior work; 6.2 on NUP-BERT (H) -- it is worth noting that NUP-BERT (TC-S) attains a ratio of 11.6, considerably worse than when trained on HUMOD data.

\end{document}